\documentclass[journal]{IEEEtran}

\usepackage{cite}
\usepackage{hhline}
\usepackage{verbatim }
\usepackage[hyperindex=true,colorlinks, citecolor=blue,menucolor=blue]{hyperref} 
\hypersetup{
	colorlinks=true, 
	citecolor=blue,
	filecolor=blue,
	linkcolor=blue,
	urlcolor=red,
	linktoc=all,     
}

\usepackage{lineno}
\usepackage{amssymb, graphicx, amsmath,  psfrag, color, array, mathrsfs, setspace, algorithm, algpseudocode, mathtools, esint}
\usepackage{amsthm}

\usepackage{pgfplots}
\usetikzlibrary{external}
\newtheorem{theorem}{Theorem}
\allowdisplaybreaks
\makeatletter
\let\MYcaption\@makecaption
\makeatother
\usepackage[font=footnotesize]{subcaption}
\makeatletter
\let\@makecaption\MYcaption
\makeatother

\begin{document}

\title{Exploiting generalization in the subspaces for faster model-based learning}

\author{Maryam~Hashemzadeh, 
         Reshad~Hosseini 
        and Majid Nili~Ahmadabadi
\thanks{The authors are with the School of ECE, College of Engineering, University of Tehran, Tehran, Iran (email: m.hashemzadeh@ut.ac.ir; reshad.hosseini@ut.ac.ir; mnili@ut.ac.ir).}
}

\maketitle

\begin{abstract}
Due to the lack of enough generalization in the state-space, common methods in Reinforcement Learning (RL) suffer from slow learning speed especially in the early learning trials.
This paper introduces a model-based method {\color{black} in discrete state-spaces} for increasing the learning speed {\color{black} in terms of required experience (but not required computational time)} by exploiting generalization in the experiences of the subspaces.  
A subspace is formed by choosing a subset of features in the original state representation (full-space). 
 Generalization and faster learning in a subspace are due to many-to-one mapping of experiences from the full-space to each state in the subspace.
 Nevertheless, due to inherent perceptual aliasing in the subspaces, the policy suggested by each subspace does not generally converge to the optimal policy. 
 Our approach, called  Model Based Learning with Subspaces (MoBLeS), calculates confidence intervals of the estimated Q-values in the full-space and in the subspaces. These confidence intervals are used in the decision making, such that the agent benefits the most from the possible generalization while avoiding from detriment of the perceptual aliasing in the subspaces.  
Convergence of MoBLeS to the optimal policy is theoretically investigated. 
Additionally, we show through several experiments that MoBLeS improves the learning speed in the early trials.
\end{abstract}



\section{Introduction}

Learning from interactions is one of the most fundamental capabilities for intelligent creatures to adapt themselves to their complex environments. Intelligent systems benefit from a variety of interactive learning methods, among those reinforcement learning (RL) has the broadest spectrum of usage (see \cite{littman2015reinforcement, sutton2016reinforcement}). RL owes its wide range of applicability mainly to its simplicity, in terms of the information content of required feedback and its low computational load, as well as to its biological implications as a cognitive model and proof of convergence to optimal policy in discrete representations. 

Nevertheless, RL severely suffers from the curse of dimensionality \cite{niv2015reinforcement}. That is, the number of interactions an RL agent requires to learn is over linearly related with the state-space size. This results in slow learning rate, high regret, and consequently limited usage in face of limited learning budget as well as in the physical systems. Inefficiency in  experience generalization over discrete state-spaces is mainly blamed for slow learning.

In RL framework, an experience is the tuple of current state, action, next state, and the received reward. Therefore, an experience generalization method targets employing experience in a limited number of states to approximate values over a larger portion of states \cite[Chapter 9]{sutton2016reinforcement}. In other words,  generalization means using experiences in some states to update values in some other states, called similar states, to attain the optimal policy with smaller number of interactions with the environment. The challenge in generalization of experiences is finding similarity across the states when the learning agent is not aware of the task model beforehand and has no high level capability to discover possible similarities during learning.  

 Some  methods have been proposed to enhance RL with the generalization capability. These methods can be clustered into seven main categories. Ensembles of RL and other learning methods, like imitation learning, and learning from other sources of information are excluded in this categorization. {\color{black} Though this paper is on discrete state-spaces, in the following we shortly describe these seven main categories that can be applied to either discrete state-spaces or continuous state-spaces.}
   
The first category of the methods generalizes experiences across multiple learning agents. In multi-agent learning \cite{stone2000multiagent, ahmadabadi2002expertness, mousavi2014context}, the agents exchange their expertise and knowledge to speed up individual learning. These methods are useful when the agents have different expertise. However, their applicability is inversely correlated with heterogeneity among the agents. In addition, these methods do not work for isolated and non-social agents. 

In the second category, generalization is embedded into state coding by the designer where similar states share some elements of the state vector \cite[Chapter 9]{sutton2016reinforcement}. This approach is the core of generalization in continuous RL; where state-action values in the neighboring states are naturally similar. {\color{black}Nevertheless, generalization through state coding can be applied in discrete environments.} 

Finding a hierarchy in the task and generalization through reusing the learned subtask as well as the structure of hierarchy are the main ideas behind the third set of methods \cite{barto2003recent}. In this category, benefiting from similarity at the subtask level is targeted.  Different algorithms have been developed to discover the hierarchy in parallel to learning (see \cite{taghizadeh2013novel} as an example).  Hierarchical methods significantly speed up the learning process and converge to the optimal policy by detecting the hierarchy, if a hierarchy exists in the task.  Nevertheless, these methods are not helpful in the early learning trials, since recognizing the hierarchy itself requires some interactions with the environment.

The fourth category of methods looks into the normalized value space, called functional space, for similarity and generalization \cite{mousavi2014context, mousavi2015context}. The main idea in employing the functional space is that, for a pair of states, similarity in the normalized values of some more experienced actions increases the probability of similarity in the Q-values of less experienced ones. Generalization through the functional space reduces regret in the early learning trials. However, there is no guarantee for convergence to the optimal policy. 

The fifth set of methods employs attention learning, that is, learning to percept the environment efficiently \cite{borji2011cost}. Here by being efficient we mean paying less cost for perception or reducing the size of state-space to learn faster. Some researchers have enabled RL to find salient features in the state vector by attention learning mechanisms.
	It is also shown that human beings employ this strategy to reduce dimensionality of their perception space \cite{niv2015reinforcement, gershman2015discovering}. When salient features are recognized, less salient features are excluded from the state definition and experience generalization emerges as the result of multiple-to-one state projection. This process is gradual and generalization shows up progressively. Attention learning itself is costly and is helpful when state features are locally, or globally, redundant.

In the sixth set of methods, efficient state representation is implicitly learned. Using  RL in deep neural networks is a major example of these methods (see \cite{mnih2015human} as an example). This approach is very promising, especially to learn complex problems. However, more research is required to benefit from it in face of limited learning budget. 

In the seventh category, our group exploited possible generalization in the subspaces to have faster learning (see \cite{firouzi2012interactive, daee2014reward}). Here a subspace refers to a sub-dimension of the original state representation, which is called the full-space. In \cite{daee2014reward}, the authors showed that learning a single step task in a multidimensional full-space can be accelerated if the agent simultaneously estimates action-value intervals in the full-space and in each subspace, and fuses the estimated values. Faster learning in \cite{daee2014reward} is due to possible generalization in some subspaces obtained because of multiple-to-one projection from the full-space into each subspace.  Therefore, every state in a subspace benefits from more experiences for action-value estimation, compared to a state in the full-space. In \cite{daee2014}, possibility of using generalization in subspaces in a Monte Carlo model-free method for multi-step tasks is explored. The results are promising; however, the method suffers from slow learning and lack of the proof of convergence. 
In this paper, we extend the idea of benefiting from possible generalization in the subspaces to speedup model-based learning of multi-step tasks, especially at the early learning trials, as well as convergence to the optimal policy.

The main contributions of this paper are summarized below:
\begin{itemize}
\item We explain the idea of using subspaces in RL framework for expediting the learning process in a concrete model. In the previous works by other members of our group, the idea of using subspaces for expediting the learning process has been exploited \cite{firouzi2012interactive, daee2014reward}. However, the idea was not exploited in a mathematically rigorous manner. 
\item This work is the first one that puts forward a model for using subspaces in the model-based RL. Previous works exploited the idea of using subspaces in the single-step and multi-step model-free RL \cite{firouzi2012interactive, daee2014reward}. 
\item We analyzed the convergence of the proposed method. There is not convergence analysis in the previous works.
\item Through several simulations, we investigate the performance of the proposed method.
\end{itemize}

The rest of paper are organized as follows. In the next section, we explain the problem and some definitions used in this paper. Next, we give an introduction of the model-based approach in RL. Afterwards, in Section~\ref{sec.proposed}, we explain the proposed model for using subspaces in model-based RL. In addition to the model, we analyze the convergence of the proposed method in this section. Next, we investigate the performance of the model in several 2D environments and 2D environments with extra sensors. Finally, we finish the paper by a discussion and giving several venues for future works.

\section{Problem statement and assumptions}
\label{sec.ProblemStatement}
The agent lives in an unknown finite discrete-time \textit{markov decision process} (MDP) with discrete state-action  defined by a tuple $\{S,A(.),p(.,.,.),R(.,.), \gamma \}$, where $S$ is a finite state-space, $A(s)$ is a finite action space in the state $s$,  $p(s,a,s')$ denotes the transition probability for $(s,a,s')$ that is the transition from the state $s$ to the state $s'$ by action $a$, $R(s,a)$  denotes the expected reward for the state-action $(s,a)$, and $0\leq\gamma\leq1$ is a discount factor for the accumulated reward~\cite{Bel}. 

{\color{black}A feature is a variable conveying some information about the environment.} Assume that the agent gets the information of all $k$ different features from the multi-dimensional environment, denoted by the feature set $\{F^1, F^2 , \ldots ,F^k\}$, where $F^i$ is the $i^{th}$ feature. Since MDP assumes fully observable agent, the feature set contains all information about the environment. Also the state-space is the Cartesian product of the features ${S = F^1\times F^2 \times \ldots \times F^k}$, called the full-space, and a state is represented by $s=(f^1,f^2,\ldots,f^k)$, where ${f^i}$ is the value of the $i^{th}$ feature.

Every non-empty subset of the full-space is called a ``feature subspace'', that we call it ``subspace'' for brevity.  So, each state in the subspaces is a projection of the state in the full-space onto the subspaces. {\color{black} Each subspace can be seen as a special case of tile coding \cite[p.~214]{sutton2016reinforcement}.}

The agent's goal is learning the optimal policy through the interaction with the environment. In this paper, the agent gradually builds the models of the task in the full-space and in some selected subspaces. Using the models, the agent approximates the optimal state-action Q-value interval in that spaces and intelligently integrates the values to increase learning speed.
We assume that the agent does not know the task model, but it knows the length of the reward interval.

\section{Background on the model-based learning}
\label{sec.background}

In the model-based approach, the model of the task is estimated from the agent experiences during the learning process. The decision-making policy is based on the estimated model and the best action computed from the estimated model. For example in $\epsilon$-greedy policy, the best action is chosen by the probability of $1-\epsilon$ and a random action by the probability of $\epsilon$~\cite{sutton2016reinforcement}.
Some approaches use confidence interval of the Q-values for their decision making policies~\cite{puterman2014markov, givan1997bounded, strehl2008analysis, ortner2007logarithmic, auer2009near}.
In this section, we give an overview of the model estimation, Q-value estimation, and confidence interval estimation of the Q-values.
\subsection{Model estimation} \label{subsection:ME}
For each transition $(s, a, s')$, the agent updates some parameters of the estimated model. These parameters are $\widehat{p}(s,a,s')$, $\widehat{R}(s, a)$ and $n(s,a,s')$, where $n(s,a,s')$ is the total number of transitions from $s$ to $s'$ by action $a$. 
These quantities are updated by the following equations:
 \begin{align}
    &\widehat{R}(s,a) \leftarrow \frac{\widehat{R}(s,a)n(s,a) + r_i(s,a) }{n(s,a)+1},\label{equ:AveReward}\\
    &\widehat{p}(s,a,s') \leftarrow \frac{n(s,a,s')+1}{n(s,a)+1},
        \label{equ:AveProb}\\
     &n(s,a,s') \leftarrow n(s,a,s')+1,
     \label{equ:num_SAS}
\end{align}
where the parameter $r_i(s,a)$ is the immediate reward of the current action and 
\begin{equation*}
n(s,a) = \sum_{s'} n(s,a,s')
\end{equation*}
is the total number of taken action $a$ in  $s$. 
\subsection{Value estimation (average, upper bound, and lower bound)} \label{subsection:VE}
The expected reward for being in a particular state, $\widehat{V}_\pi(s)$, and the expected reward of taking action $a$ in state $s$, $\widehat{Q}_\pi(s,a)$, by exploiting the policy $\pi$ are given by the following Bellman equation \cite{Bel}:
 \begin{align}
      \widehat{Q}_\pi(s,a)\;&=\; \widehat{R}(s,a)+ \sum_{s'}  \widehat{p}(s,a,s')\left[ \gamma \widehat{V}_\pi(s')\right], 
      \label{equ:Q_Aprox}\\
       \widehat{V}_\pi(s)\;&=\; \max_a \widehat{Q}_\pi(s,a).
       \label{equ:V_Aprox}
\end{align}
 A commonly used method for computing the optimal expected rewards, $\widehat{Q}^*(s,a)$, of the estimated model is \textit{policy evaluation} (PE) algorithm  \cite{ Bel,howard1960dynamic}. 

If the exact values of the model parameters are known, the best policy can  be calculated by simply taking the action that maximizes the optimal Q-values, $Q^*(s,a)$, for the exact model. 

In the model-based approaches, the model parameters are estimated by the previous experiences and contain uncertainty; therefore, the estimated Q-values, $\widehat{Q}^*(s,a)$, have uncertainty too. Confidence intervals can be used to model the uncertainty of the estimated parameters \cite{puterman2014markov, givan1997bounded}.

The confidence intervals of the (estimated) Q-values are computed in two steps \cite{givan1997bounded}:
\begin{enumerate}
\item{Calculating the confidence intervals for the estimated parameters.}
\item{Calculating the upper and lower bounds on the Q-values from the confidence intervals.}
\end{enumerate}
\subsubsection*{Step 1- Calculating the confidence interval of the estimated parameters}
In the following, we first explain how the confidence intervals of the estimated rewards are calculated. Then, we give the expression for the confidence intervals of the estimated transition probabilities in the next part. 
	\paragraph{Confidence intervals of the estimated rewards}$ $
	
We assume that the agent does not know the distribution of immediate rewards. We only know that immediate rewards are samples from random variables defined in a fixed interval and they are independent. We choose to use the Hoeffding inequality \cite{hoeffding1963probability} that is distribution-free and can be used to compute the confidence intervals. Hoeffding inequality is expressed by the following equation:
\begin{align}
p & \left( \vert \widehat{R}(s,a) -R(s,a ) \vert <  \varepsilon_R\right) \geq  \nonumber \\
& 1- 2\exp \left(⁡{-\frac{2n(s,a)^2 \varepsilon_R^2}{\sum_{i=1}^{n(s,a)} (\beta_i(s,a) -\alpha_i(s,a))^2}}\right),
\label{eq:EqualityforReward}
\end{align} 
where $\varepsilon_R$ is an arbitrary number and $[\alpha_i(s,a), \beta_i(s,a)]$ is an interval, that $i$th immediate reward for the state-action $(s, a)$ is sample from a random variable in it.
{\color{black} Consider the right side of~\eqref{eq:EqualityforReward} that is called confidence coefficient is equal to $1-\delta_R$.}  Then, the value of $\varepsilon_R $ can be computed by the following relation:
 \begin{equation}
\varepsilon _ R \;=\; \sqrt{\frac{\sum _{i=1}^{n(s, a)} (\beta_i(s,a) - \alpha_i(s,a))^2 \log \left({\frac{2}{\delta_R}}\right)}{2n(s, a)^2}}.
\label{eq:epsilon_R}
\end{equation}

%
%
\paragraph{Confidence interval of the estimated transition probabilities}$ $

We assume that the environment is stationary and therefore the actual transition probabilities are fixed. The estimated transition probabilities are given by~\eqref{equ:AveProb} and we can use Weissman inequality \cite{weissman2003inequalities} for computing the confidence intervals.

Consider $P(s, a)$ to be a probability vector having the transition probabilities from $s$ to other states by taking the action $a$ as its elements. A probability vector is a vector with positive elements that its elements sum  to one.
 The Weissman inequality for the probability vector is expressed by the following equation:
 \begin{align}
 p & \left\{ \Vert \widehat{P}(s,a)-P(s,a)\Vert _1 \leq \varepsilon _p \right \} \geq  \nonumber \\
& 1-(2^{m(s)} -2) \exp \left(-\frac{n(s,a)\varepsilon_p^2}{2}  \right),
\label{eq:Wiss}
\end{align}
where $\widehat{P}(s,a)$ denotes the estimated transition probability vector, $m(s)$ is the dimensionality of the probability vector $P(s, a)$ and $\varepsilon_p$ is an arbitrary number. Consider the right-hand side of  the relation \eqref{eq:Wiss} is equal to $1-\delta_p$, where $1-\delta_p$ is the confidence coefficient for the transition probabilities. Then,  the value of $\varepsilon_p$ is equal to
  \begin{equation}
\varepsilon_p = \sqrt{\frac{2\log \left(\frac{2^{m(s)} - 2}{\delta_p}\right)}{\max \{1, n(s, a)\}}}.
\label{eq:EpsilonP}
\end{equation}
The confidence interval of the estimated transition probability is defined by:
 \begin{equation}
CI_P = \left\{\widetilde{P}(s,a)\in P  \, \vert \; \Vert \widetilde{P}(s,a)- \widehat{P}(s,a)\Vert _1 \leq \varepsilon _p\right\},
\label{eq:CI_p}
\end{equation}
where $P$ is the set of all $m(s)$-dimensional probability vectors.  

It is clear that by gaining more experience, i.e. increasing $n(s, a)$, the confidence intervals for $\widehat{P}(s,a)$ and  $\widehat{R}(s,a)$ become smaller.

\subsubsection*{Step 2- Calculating the upper and lower bounds on the Q-values}
 The upper and lower bounds on the Q-values can be seen as the optimistic and pessimistic expected rewards for each state-action pairs. We denote these upper and lower bounds by $Q_u(s, a)$ and $Q_l(s,a)$, respectively.
By making use of the confidence intervals on $\widehat{P}(s,a)$ and  $\widehat{R}(s,a)$, the optimistic and pessimistic Q-values of each state-action pairs are calculated.  Algorithm \ref{alg:UB}  illustrates the procedure of computing $Q_u(s, a)$ and  $Q_l(s,a)$. This algorithm is strongly related to the \textit{internal policy evaluation} given in Givan et al.~\cite{givan1997bounded} and  the same as the \textit{extended value iteration } in MBIE and UCRL2 algorithms \cite{strehl2008analysis, auer2009near}. The only difference is that  we do not compute the optimistic policy after computing the confidence interval for Q-values.

Algorithm \ref{alg:UB} solves the following optimization problems for computing the optimistic and pessimistic Q-values:
\begin{align}
Q_u(s, a)&=\left( \widehat{R}(s, a)+ \varepsilon_R \right)  +  \nonumber \\
 &\max_{\widetilde{P}(s, a)\in CI_p} \sum _{s'} \widetilde{p}(s, a, s') \left(\gamma  \max _{a'}\widehat{Q}_u(s', a') \right),
\label{eq:Q_u}\\
Q_l(s, a)&= \left( \widehat{R}(s, a)- \varepsilon_R \right)  + \nonumber \\
&\min_{\widetilde{P}(s, a)\in CI_p} \sum _{s'} \widetilde{p}(s, a, s') \left( \gamma  \max _{a'}\widehat{Q}_l(s', a') \right),
\label{eq:Q_l}
\end{align}
where $\widetilde{p}(s, a, s')$ is an element of the probability vector $\widetilde{P}(s, a)$.

\renewcommand{\algorithmicrequire}{\textbf{Input:}}
\begin{algorithm*}[!ht]
\newcommand{\NewComment}[1]{ {\hfill$//$ #1}} 
  \caption{Calculating the optimistic and pessimistic Q-values for each state-action pairs based on the estimated model}
  \label{alg:UB}
	    \selectfont
  	    \footnotesize {\textbf{Function} OptimisticValue($ n,\widehat{p}, \widehat{R}, \text{LenR}, \delta_R, \delta_p, \gamma, \theta$)}
		\NewComment{ \parbox[t][]{\dimexpr.4\linewidth-\algorithmicindent}
			    {For calculating the  pessimistic Q-values, replace the commented phrases.}}
    \begin{algorithmic}[1]
	    \Require 
 	    {$n$ stores counters,  $\widehat{p}$ is the estimated probabilities vector, $\widehat{R}$ is the estimated rewards vector,  $\text{LenR}$ is the vector containing the length of rewards,  parameters $1-\delta_R \in [0, 1]$ and  $1-\delta_p \in [0, 1]$ are the confidence coefficients,  $\gamma$ is the discount factor,    $\theta$ is a small positive number used in the stopping criterion}\\ 
 	   
	    \footnotesize{\textbf{Initialize}\; $Q_u(., .)=0,  M= - \infty$}\nonumber
		\NewComment{ \parbox[t][]{\dimexpr.3\linewidth-\algorithmicindent}{$Q_l(., .)=0$}}
	    \Repeat 
	    \ForAll{$s \in S$}
	    \ForAll{$a \in A$}
	    \State $q=Q_u(s, a)$
		\NewComment{ \parbox[t][]{\dimexpr.3\linewidth-\algorithmicindent}{$q=Q_l(s, a)$}}
	   \State $\varepsilon _ R \;=\; \sqrt{\frac{\sum _{i=1}^{n(s, a)} \text{LenR}(s, a)^2 \log \left({\frac{2}{\delta_R}}\right)}{2n(s, a)^2}}$
	    \State  $R_u(s,a) = \widehat{R}(s, a) + \varepsilon_R $
		\NewComment{ \parbox[t][]{\dimexpr.3\linewidth-\algorithmicindent}{$R_l(s,a) = \widehat{R}(s, a) - \varepsilon_R $}}
		 \State $k=0$
		\ForAll{$s' \in (\text{states reachable from state } s \text{ by taking action }a) $}
 \Comment{ \parbox[t][]{\dimexpr.3\linewidth-\algorithmicindent}{If it is not known, it is set to all states $S$.}}
		    \State $k\leftarrow k+1$
		    \State $\psi(k) = s'$
		    \State  $ \widehat{p}(s, a, s')= \dfrac{n(s, a, s')}{\sum _ {s'}n(s, a, s')}$
		    \State  $ U(k)=  \gamma \max _{a'} Q_u(s', a')$
			\NewComment{ \parbox[t][]{\dimexpr.3\linewidth-\algorithmicindent}{$ U(k) =  \gamma \max _{a'} Q_l(s', a')$}}
		\EndFor

		\State  $idx =$ Indices of sorted $U$ in decreasing order
		    \NewComment{ \parbox[t][]{\dimexpr.3\linewidth-\algorithmicindent}{$idx =$ Indices of sorted $U$ in increasing order}}
		\State $i=idx(1)$ 
	\State $m(s)=\vert U \vert$
	\State $\varepsilon_p = \sqrt{\frac{2\log \left(\frac{2^{m(s)} - 2}{\delta_p}\right)}{\max \{1, n(s, a)\}}}$
	\State $\widehat{p}(s,a, \psi(i))=\min \left(1, \widehat{p}(s,a,\psi(i))+\dfrac{\varepsilon_p}{2} \right)$
		\State $j= \vert idx \vert +1$
		\Repeat

		\State $j=j-1$
		\State $i=idx(j)$
		\State $ P=\sum_{k \in idx} \widehat{p}(s, a,\psi(k)) - \widehat{p}(s, a, \psi(i))$
		\State $\widehat{p}(s,a, \psi(i))=\max (0, 1-P)$
		\Until{$P+ \widehat{p}(s, a, \psi(i)) > 1$}
		\State $Q_u(s, a)= R_u(s,a) + \text{Inner product}\left(\widehat{p}(s,a,.), U(.) \right)$
		    \NewComment{ \parbox[t][]{\dimexpr.4\linewidth-\algorithmicindent}{ $Q_l(s, a)= R_l(s,a) + \text{Inner product}\left(\widehat{p}(s,a, .), U(.)\right)$}}
		\State $Dif = \vert q-Q_u(s, a) \vert$ 
		    \NewComment{ \parbox[t][]{\dimexpr.3\linewidth-\algorithmicindent}{$ Dif = \vert q-Q_l(s, a) \vert$}}
			\If {$Dif>M$}
		\State $M=Dif$
		\EndIf

	    \EndFor
	    \EndFor
	    \Until{$M >\theta $}\\
	    \Return{$Q_u$}
	    \NewComment{ \parbox[t][.5cm]{\dimexpr.3\linewidth-\algorithmicindent}{$Q_l$}}
	\end{algorithmic}
\end{algorithm*}

\section{Proposed Method}
\label{sec.proposed}
In our methodology, we use  generalization of the agent experiences in the feature subspaces to speed up the learning process. Since the subspaces are low-dimensional, learning in them needs smaller number of experiences in compare to the full-space. 

\begin{figure}
    \centerline{
	      \includegraphics[width=0.5\textwidth]{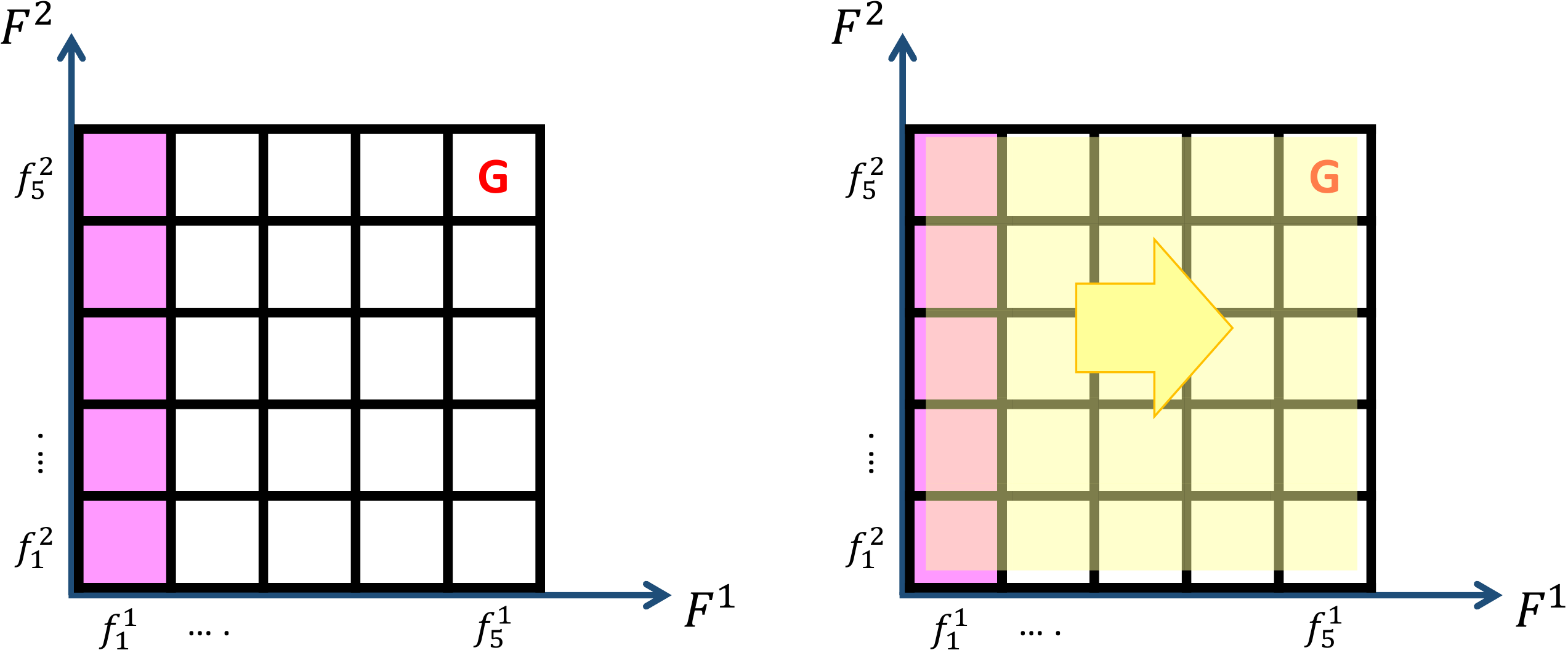}}
     \caption{An example showing a $5\times 5$ maze. Columns and rows can be both two feature sets and two subspaces. The values of the feature sets are $\{ 1, 2, 3, 4, 5\}$. The agent can do four actions, either going up, down, right, or left, to reach the goal specified by $G$. Due to the goal location, if the agent learns that action right is preferable in each column, the learning speed can increase. Therefore, the generalization of the experiences in the subspaces can increase the learning speed.}
    \label{fig:maze_X}
    \end{figure}

Fig. \ref{fig:maze_X} illustrates an example of a $5\times 5$ maze to demonstrate how generalization in the subspaces can speed up the learning process. The environmental features are   $\{F^1\}$ and $\{F^2\}$ that correspond to x and y positions, respectively. Its full-space is $\{ F^1, F^2\}$ and the state-space is $S= F^1\times  F^2$. The subspaces are $\{F^1\}$, $\{F^2\}$.  The agent can do four actions, either going up, down, right, or left, to reach the goal specified by $G$.  Due to the goal location, if the agent learns that action right is preferable in each column, the learning speed can increase. Suppose the agent has already gone in state $(1, 2)$ for $5$ times and in state $(1, 3)$ for $4$ times. If the agent goes to state $(1,4)$ for the  first time, it will choose an action randomly since there is no full-space experience in this state. 
If the agent relies also on the subspace experiences, it may choose the right action because the agent has been in the first column for 9 times. This, we call generalization of the experiences in the subspaces. Consequently, the learning process would be faster if the agent relies more on the subspace experiences in the beginning of the learning process.


The generalization of the agent experiences is particularly helpful when some states in the full-space are rarely visited while the corresponding states in the subspaces are visited more often. For example, assume the agent goes to a new state in the full-space while has been in the corresponding state of the subspaces repeatedly. If the agent wants to decide appropriate action based on the full-space experiences, it nearly chooses a random action. However, the agent can utilize the generalization of the experiences in the subspaces to further speed up learning.

There is also the problem of \textit{perceptual aliasing} (PA) in the subspaces. This is when the environment is partially observable in the perspective of the subspaces; thus, there is a many-to-one mapping between a state of the full-space and a state of a subspace~\cite{whitehead1991learning}. 
Due to PA, the decisions in the subspaces can be suboptimal, because the agent does not see all environmental features. For example in the previous maze example, assume there are some barriers in some positions in the second column. The optimal policy in those positions in the first column, where the adjacent position in the second column is a barrier is not the decision of going to the right. Therefore, making appropriate use of the generalization  needs to evaluate its usefulness respect to the PA issue.

Indeed, for using the subspaces experiences in the learning process, there are two influential factors, the profit of the generalization and the detriment of the PA issue. 
Assume the agent can select between the decision suggested by the full-space and the subspaces.
 At the early stages of the learning process, because the agent has less experience, the decision of the full-space is unreliable. The agent should rely more on the subspace decisions, because the profit of the generalization is more than the detriment of the PA issue. By gaining more experience, the uncertainty in the estimated model of the full-space becomes less and the agent should choose the full-space decision, because the loss of the PA issue in the subspaces increases.

 {\color{black}It is obvious that the task for each subspace also satisfies the Markov property, therefore the reinforcement learning tasks in the subspaces are also MDP.}
  In the proposed approach for increasing the learning speed, the policy suggested by the full-space is combined with the subspaces' policies of the learned MDPs.
Because we are using the model-based approach, the task parameters are estimated during the learning process in both the full-space and subspaces in accordance with the environmental feedbacks. So as to reduce the effect of the PA issue, uncertainty of each suggestion for decision-making is calculated by the confidence interval of the estimated task parameters. 

\begin{figure}
    \centerline{
    \includegraphics[width=1\columnwidth]{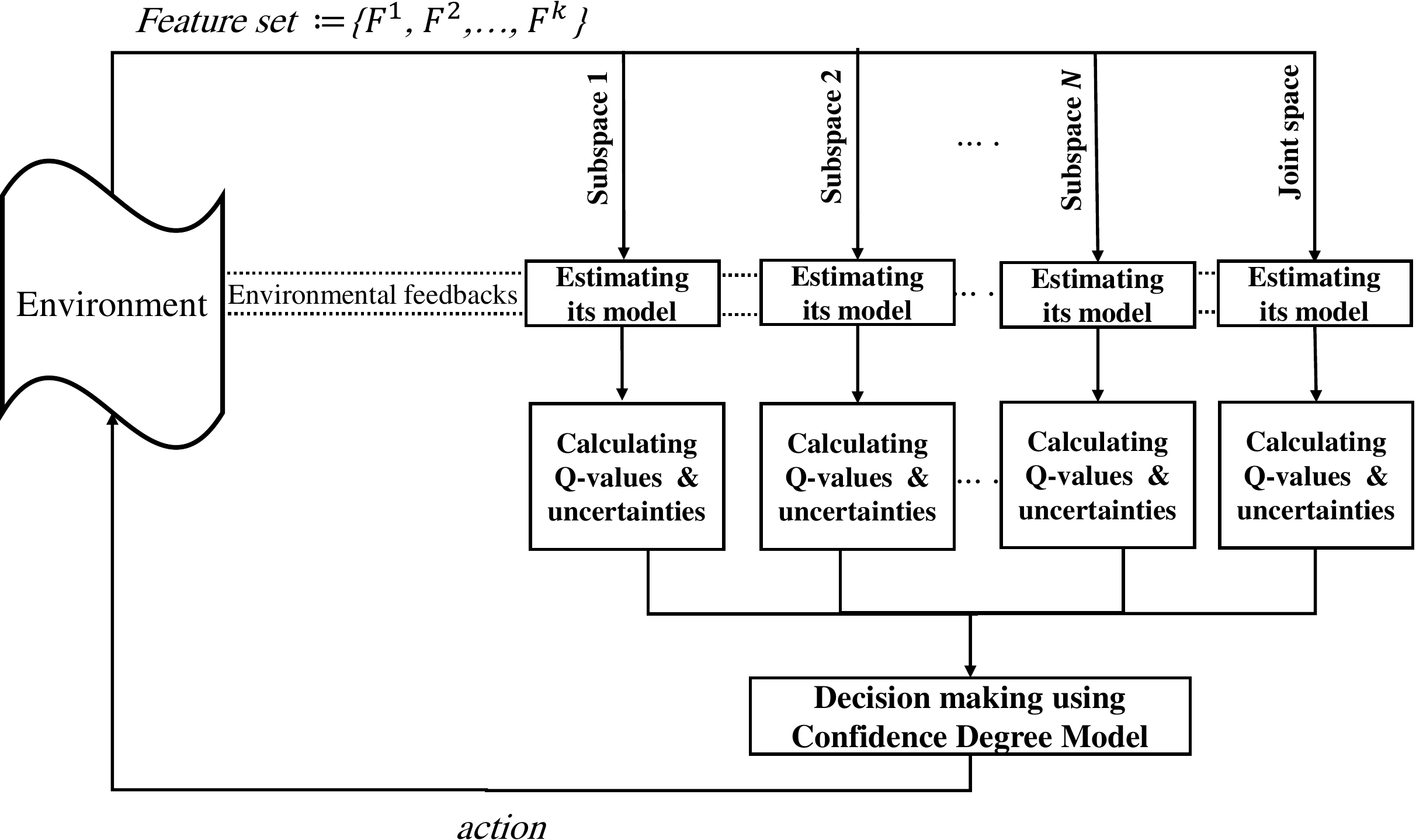}}
\caption{A schematic overview of the proposed framework for using the subspaces experiences to speed up the learning process. The feature set are $\{F^1, F^2 , \ldots ,F^k\}$. The agent updates the model parameters of the full-space and subspaces by using environmental feedbacks to its actions. In accordance with the estimated model of each space, the Q-values and their uncertainties for all state-action pairs are calculated in  the subspaces and the full-space. Finally, the agent integrates all information to make its decision by employing CDM.}
\label{fig:Flowchart2}    
\end{figure}

The scheme of the proposed method is summarized in Fig. \ref{fig:Flowchart2}. The proposed method comprises three main phases that is performed recursively: 1) estimating the subspaces and full-space models, 2) calculating the Q-values and uncertainties, and 3) combining the decisions of the subspaces and the full-space to make the decision in the current state. Based on the Q-values and uncertainties, a weight  is assigned to each action in the subspaces and the full-space. In the last stage, a model is used to integrate the acquired choice probabilities for choosing  appropriate action. We call this model \textit{confidence degree model} (CDM). In the following subsections, the proposed  learning and decision making method is explained in details.

\subsection{Confidence interval estimation}
We use confidence intervals to model the uncertainty on the estimates for the full-space and subspaces. We use these confidence intervals in CDM, so to make the most benefit of the generalization in the subspaces while taking care of the PA issue. 

Confidence intervals are estimated for the full-space and subspaces based on the estimated parameters of MDPs for each of them. The model parameters estimation procedure and value estimation were already explained in Subsection \ref{subsection:ME} and~\ref{subsection:VE}. Since the PE algorithm is computationally demanding, we run the algorithm \textbf{only after each episode} to update the optimal expected rewards.

For computing the confidence interval of the Q-values for the full-space and subspaces, we need to calculate $\varepsilon_R$ and $\varepsilon_P$ according to equations \eqref{eq:epsilon_R} and \eqref{eq:EpsilonP} for each space. Assume all immediate rewards {\color{black}in different learning steps for a specific} state-action $(s_j, a)$ in the full-space have the same interval denoted by
$[\alpha_{\text{fspace}}(s_j,a)\ ,  \beta_{\text{fspace}}(s_j,a)]$. Then, for the full-space $\varepsilon _ R$ is given by:
  \begin{equation}
 \varepsilon _R \; = \; \sqrt{\frac{(\beta_{\text{fspace}}(s_j, a)-\alpha_{\text{fspace}}(s_j, a))^2 \log \left(\frac{2}{\delta_R} \right)}{2n(s_j, a)}}.
 \label{eq:NewEpsilonRJoint}
 \end{equation}
For the subspaces $\varepsilon _ R$ is given by:
 \begin{equation}
\varepsilon _ R \;=\; \sqrt{\frac{\sum _{i=1}^{n(s_x, a)} (\beta_i(s_x, a) - \alpha_i(s_x, a))^2 \log \left({\frac{2}{\delta_R}}\right)}{2n(s_x, a)^2}},
\label{eq:NewEpsilonRSub}
\end{equation}
where $s_x$ and $s_x'$ are the states in the subspace $x$. Note that for the subspaces, the interval depends on the immediate reward at the $i$th observation. This is because there is a many-to-one mapping from the states in the full-space to a state in the subspace. Therefore, every reward sample for each transition in the subspaces can have different intervals.

\subsection{Combining policies of the full-space and subspaces}\label{sec:CDM}
In Section~\ref{sec.background}, we explained how to estimate Q-values and their uncertainties for every state-action pairs in the full-space and the subspaces. 
We use a soft decision policy based on estimated Q-values, so to have both exploration and exploitation. 
We use available techniques in the literature for computing policies for the full-space and subspaces separately. Meaning that a subspace or the full-space is considered a separate MDP and a decision policy is computed for it. For computing the final policy for each state-action pair, we use the CDM to integrate these decision policies.

For integrating decision policies, we assign a  coefficient to each subspace and then based on these coefficients the final policy is calculated. Let us first discuss how to calculate these coefficients, we will come to the procedure for combining these later. 
The coefficients, that we call them \textit{confidence degree}s, are estimated from the estimated Q-values and their confidence intervals. {\color{black} If the confidence interval of a state-action pair for a subspace does not overlap with the confidence interval of its corresponding full-space, then the confidence degree of that subspace is set to zero. }
Otherwise, the confidence degrees for the subspaces are computed by multiplying following two quantities:
\begin{enumerate}
\item{Length quantity
}
\item{Overlap-distance quantity
}
\end{enumerate} 
In the following subsections, we explain each of them in details. 
\subsubsection{Length Quantity} \label{SubSection:LengthQuantity}
When the confidence intervals are larger, the estimated Q-values have larger uncertainty. 
The length quantity that computes the effect of the uncertainty of each subspace $s_x$ is computed by the following equation: 
\begin{equation}
\label{eq:fx}
f\left(\frac{\vert CI(s_j, a)\vert}{\vert CI(s_x, a)\vert} \right),
\end{equation}
where $\vert CI(s_x, a)\vert$ is the length of the confidence interval $CI(s_x, a)$ for each $(s_x, a)$ in each subspace, and $\vert CI(s_j, a)\vert$ is the length of the confidence interval for the corresponding full-space state-action. Function $f$ is an increasing function determined based on experiments. It is chosen to be the following function in our simulations.
\begin{equation}
\label{fx2}
f (t) =
  \begin{cases}
   0       & t \leq1\\
    t  & t>1\\
  \end{cases}.
\end{equation}

In the early stages of the learning process, the experiences in the subspaces are significantly more than the experiences in the full-space because several state-action pairs in the full-space map to one state-action pair in the subspace.
Therefore, the uncertainty in the subspaces are less than the full-space and the agent should rely more on the subspaces decisions. Thus, the confidence degrees should be higher for the subspaces decisions and the length quantity given by~\eqref{eq:fx} takes care of this.

\subsubsection{Overlap-distance Quantity}
\label{SubSection:Overlap-distance-Quantity}
This quantity measures the consistency between experiences in a subspace with its corresponding experiences in the full-space. If the experience of a subspace is more consistent with its corresponding experiences in the full-space, it means that the PA issue is less problematic in this state and the agent can rely more on the subspace decision. The overlap-distance quantity is computed by the following relation:
\begin{equation}
\label{gx}
g \left(\frac{\vert \widehat{Q}_\pi (s_j,a) - \widehat{Q}_\pi (s_x, a)\vert}{\vert CI(s_j, a)\cap CI(s_x, a) \vert}\right),
\end{equation}
where $\widehat{Q}_\pi(s_x,a)$ and $\widehat{Q}_\pi(s_j, a)$ are the estimated Q-value of  the state-action pair $(s_x,a)$ in subspace $x$ and its corresponding value in the full-space, respectively. Function $g$ is a decreasing function and for our experiments is chosen to be the following function:
\begin{equation}
\label{gx2}
g (t) =
  \begin{cases}
   \text{-erfinv}\left(2t-1 \right)       & t \leq1/2\\
    0  & t>1/2\\
  \end{cases}.
\end{equation}
The numerator inside the function in~\eqref{gx} takes into account the distance of the estimated Q-value in the subspace and the corresponding value in the full-space.
The denominator shows the overlap between  the confidence interval of  the subspace state-action pair $(s_x, a)$ and its corresponding full-space state-action pair $(s_j, a)$. Small overlap means that the experiences in the subspace are more consistent with other states of the full-space that map to $s_x$ and not $s_j$. If two subspaces have equal overlap (i.e. equal denominator) but the distance between their estimated Q-values to the estimated Q-value of the full-space is different, the numerator ensures that the overlap-distance quantity puts more weight to that subspace that has less distance.

The overlap-distance quantity handles the problem of the PA issue in the subspaces. By gaining more experience, the full-space decision becomes reliable and the agent should rely more on it. 
In the limit, the estimated Q-value in the full-space converges to one point and the estimated Q-values of the subspaces converge to other points. So the length of the overlap reduces compared to its interval length. During the learning process, the distance between the estimated Q-values of the  subspaces and the full-space grows. Therefore, the influence of  the subspaces  reduces gradually. 

{\color{black}
Consequently, the confidence degree $CD(s_x, a)$ for $(s_x, a)$ is given by the following equation if $CI(s_j, a)\cap CI(s_x, a) \neq \emptyset$; otherwise, $CD(s_x, a)$ is set to zero. The procedure to compute confidence degrees is also summarized in Algorithm \ref{alg:CDM}.
 \begin{equation}\small
\label{eq:DC_xa}
CD(s_x, a) = f\left(\frac{\vert CI(s_j, a)\vert}{\vert CI(s_x, a)\vert} \right)  \times g \left(\frac{\vert \widehat{Q}_\pi (s_j,a) - \widehat{Q}_\pi (s_x, a)\vert}{\vert CI(s_j, a)\cap CI(s_x, a) \vert}\right)
\end{equation}}

Now, we are in the state to explain how decision integration is performed based on the confidence degrees. Consider $\alpha(s_j,a)$ to be the subspace with highest confidence degree, i.e., $\alpha(s_j,a) = \arg \max_ {x \in S_j} CD(s_x,a)$, where $S_j$ is the set of the subspaces. The final {\color{black} unnormalized} probability of the action selection in state $s_j$ is given by:
 \begin{equation}
Pr(s_j, a)
= 
\begin{cases}
p_{\pi}(s_{\alpha(s_j,a)},a)   & \quad \text{if } CD(s_{\alpha(s_j,a)},a)>1\\
p_{\pi}(s_j,a) & \quad \text{if } CD(s_{\alpha(s_j,a)},a)\leq1,\\
\end{cases}
\label{eq.int}
\end{equation}
where the value $p_{\pi}(s_x, a)$ shows the choice probability of the  action $a$ in  $s_x $  using the policy $\pi$ (for example $\epsilon$-greedy) that is computed by the model parameters of that subspace MDP. 
The value $p_{\pi}(s_j, a)$ is the choice probability in the corresponding full-space MDP. 

 The summary of our proposed method is illustrated in Algorithm~\ref{alg:CIMB}. In the following, we provide the proof that in the limit of the large number of the experiences, the confidence degrees of the subspaces converge to zero and therefore the decision-making is purely based on the full-space decision. Therefore, it further elucidates that the proposed method of choosing confidence degrees takes into account the problem of the PA issue in the subspaces.

\begin{theorem}
Let M be an MDP with finite state-space S, finite action space A
and consider the decision-making procedure is based on CDM. If the number of the experiences {\color{black}for all state-actions} goes to infinity, the Q-values of the subspaces converge. Furthermore, if the Q-values of a subspace and its corresponding full-space don't converge to the same number, then the confidence degree of that subspace converges to zero.
\end{theorem}

\begin{proof}
%
%



In the limit of infinite number of the experiences based on the \textit{low of large number} \cite{bertsekas2002introduction}, the estimated transition probability and the expected reward of the full-space converges to their actual values. Therefore, the decision policy of the full-space converges to the optimal policy. We will also show that the estimated $p$ and $R$ for 
the subspaces converge. Therefore, confidence intervals $CI_P$ and $CI_R$ converge to zero and subsequently $CI$s of the Q-values converges to zero. It is evident from~\eqref{eq:DC_xa} that when $CI$s converge to zero and the Q-values of the full-space and a subspace converge to different quantities, then the confidence degree of that subspace converges to zero.


We split the whole feature set $S$ to the subspace feature $x$ and its complement $x^c$. Then each state of the full-space $s_j$ can be represented by the vector $[s_x, s_{x^c}]$. We want to compute the estimated probability of the transition $(s_x,a,s'_x)$. We write $s_x^i$ as $i$th state reachable from state $s_x$ by taking action $a$. 

 

{
{
\setlength{\arraycolsep}{4pt}
\def\arraystretch{.5}
\begin{align}
\label{eq:proof_prob}
	\widehat{p}( & s_x,a ,s_x')  = \frac{\sum_{s_{x^c}}{ \sum_{s_{x^c}'} n \left( \begin{bmatrix} s_x\\ s_{x^c} \end{bmatrix} , a,  \begin{bmatrix} s_x' \\ s_{x^c}'  \end{bmatrix} \right)}}{{\sum_{s_{x^c}}{ \sum_ {s_{x^c}'} {\sum _{s_x^i}n \left( \begin{bmatrix}s_ x \\ s_{x^c}  \end{bmatrix} , a,  \begin{bmatrix} s_x^i \\ s_{x^c}'  \end{bmatrix} \right)}}}} \nonumber \\[1em] 
	& = \frac{{\sum_{s_{x^c}}{ \sum_ {s_{x^c}'} {n \left( \begin{bmatrix} s_x \\ s_{x^c}  \end{bmatrix}\right) Pr\left( \begin{bmatrix} s_x \\ s_{x^c}  \end{bmatrix}, a\right) \widehat{p} \left(\begin{bmatrix} s_x \\ s_{x^c}  \end{bmatrix}, a,  \begin{bmatrix} s_x' \\ s_{x^c}' \end{bmatrix} \right)}}}}
	{\sum_{s_{x^c}}{ \sum_{s_{x^c}'}{\frac{n \left( \begin{bmatrix} s_x \\ s_{x^c}  \end{bmatrix} , a,  \begin{bmatrix} s_x' \\ s_{x^c}'  \end{bmatrix} \right)}{\widehat{p} \left( \begin{bmatrix} s_x \\ s_{x^c}  \end{bmatrix} , a,  \begin{bmatrix} s_x' \\ s_{x^c}'  \end{bmatrix} \right)} }}} \nonumber \\[1em] 
	& =  \frac{{\sum_{s_{x^c}}{ \sum_ {s_{x^c}'} {n \left( \begin{bmatrix} s_x \\ s_{x^c}  \end{bmatrix}\right) Pr\left( \begin{bmatrix} s_x \\ s_{x^c}  \end{bmatrix}, a\right) \widehat{p} \left(\begin{bmatrix} s_x \\ s_{x^c}  \end{bmatrix}, a,  \begin{bmatrix} s_x' \\ s_{x^c}'  \end{bmatrix} \right)}}}}
	{\sum_{s_{x^c}}{ \sum_{s_{x^c}'}{\frac{{{  {n \left( \begin{bmatrix} s_x \\ s_{x^c}  \end{bmatrix}\right) Pr\left( \begin{bmatrix} s_x \\ s_{x^c}  \end{bmatrix}, a\right) \widehat{p} \left(\begin{bmatrix} s_x \\ s_{x^c}  \end{bmatrix}, a,  \begin{bmatrix} s_x' \\ s_{x^c}'  \end{bmatrix} \right)}}}}{\widehat{p} \left( \begin{bmatrix} s_x \\ s_{x^c}  \end{bmatrix} , a,  \begin{bmatrix} s_x' \\ s_{x^c}' \end{bmatrix} \right)} }}} \nonumber \\[1em] 
	& =  \frac{{\sum_{s_{x^c}}{ \sum_ {s_{x^c}'} {n \left( \begin{bmatrix} s_x \\ s_{x^c}  \end{bmatrix}\right) Pr\left( \begin{bmatrix} s_x \\ s_{x^c}  \end{bmatrix}, a\right) \widehat{p} \left(\begin{bmatrix} s_x \\ s_{x^c}  \end{bmatrix}, a,  \begin{bmatrix} s_x' \\ s_{x^c}'  \end{bmatrix} \right)}}}} {\sum_{s_{x^c}}{ \sum_ {s_{x^c}'} {n \left( \begin{bmatrix} s_x \\ s_{x^c}  \end{bmatrix}\right) Pr\left( \begin{bmatrix} s_x \\ s_{x^c}  \end{bmatrix}, a\right)}}},
\end{align}
}}
where $n([s_x \; s_{x^c}]^T)$ is the number of the times that the agent has visited the state $s_j$, $Pr([s_x\; s_{x^c}]^T, a)$ defined in~\eqref{eq.int}. 
The convergence of $\widehat{p}(s_x,a,s_x')$ depends on the convergence of $\widehat{p} ([s_x \; s_{x^c}]^T, a, [s_x' \; s_{x^c}']^T)$ and  $Pr([s_x \; s_{x^c}]^T, a)$. 

We prove the convergence of the subspace transition probability by contradiction.
Assume $\widehat{p}(s_x,a,s_x')$ doesn't converge to $p(s_x,a,s_x')$ and based on the algorithm for computing the confidence intervals, that is Algorithm \ref{alg:UB}, $\vert CI(s_x, a)\vert$ remains greater than zero. Since  $\vert CI(s_j, a)\vert$ converges to zero, therefore based on~\eqref{eq:DC_xa} we conclude $CD(s_x, a)\to0$ and consequently the probability action selection converges to that of the full-space. Thus, we conclude that $\widehat{p}(s_x,a,s_x')$ converges which contradicts the assumption. 

In the following, we give an expression for the estimated reward of the state-action pair $(s_x,a)$,
{
{
\setlength{\arraycolsep}{4pt}
\def\arraystretch{.5}
\begin{align}
\label{eq:proof_rew}
\widehat{R}(s_x,a) & = \frac{\sum_{s_{x^c}}{ {\sum_ {i=1}^ {n([s_x,s_{x^c}]^T,a)} r_i \left( \begin{bmatrix} s_x \\ s_{x^c} \end{bmatrix} , a \right)}}}{{\sum_{s_{x^c}}{ \sum_ {s_{x^c}'} {\sum _{s_x^i}n \left( \begin{bmatrix}s_ x \\ s_{x^c}  \end{bmatrix} , a,  \begin{bmatrix} s_x^i \\ s_{x^c}'  \end{bmatrix} \right)}}}} \nonumber \\[1em]
& = \frac{{\sum_{s_{x^c}}{ \sum_ {s_{x^c}'} {n \left( \begin{bmatrix} s_x\\ s_{x^c}  \end{bmatrix}\right) Pr\left( \begin{bmatrix} s_x \\ s_{x^c}  \end{bmatrix}, a\right) \widehat{R} \left(\begin{bmatrix} s_x \\ s_{x^c}  \end{bmatrix}, a \right)}}}}{{\sum_{s_{x^c}}{ \sum_ {s_{x^c}'} {\sum _{s_x^i}n \left( \begin{bmatrix}s_ x \\ s_{x^c}  \end{bmatrix} , a,  \begin{bmatrix} s_x^i \\ s_{x^c}'  \end{bmatrix} \right)}}}} \nonumber \\[1em]
&= \frac{{\sum_{s_{x^c}}{ \sum_ {s_{x^c}'} {n \left( \begin{bmatrix} s_x \\ s_{x^c}  \end{bmatrix}\right) Pr\left( \begin{bmatrix} s_x \\ s_{x^c}  \end{bmatrix}, a\right) \widehat{R} \left(\begin{bmatrix} s_x \\ s_{x^c}  \end{bmatrix}, a \right)}}}}
{\sum_{s_{x^c}}{ \sum_{s_{x^c}'}{{{{  {n \left( \begin{bmatrix} s_x \\ s_{x^c}  \end{bmatrix}\right) Pr\left( \begin{bmatrix} s_x \\ s_{x^c} \end{bmatrix}, a\right) }}}}}}}.
\end{align}
}}
Similarly, we can use the principle of contradiction to show that $\widehat{R}(s_x, a)$ converges. Consequently, the Q-value of the subspace also converges. 

Since the confidence intervals go to zero and we assumed that Q-values converge to different numbers, $g$ in~\eqref{eq:DC_xa} converges to zero and therefore $CD(s_x, a)\to0$.
\end{proof}

\renewcommand{\algorithmicrequire}{\textbf{Input:}}
\begin{algorithm*}[!ht]
  \caption{Confidece Degree Model for decision-making}
  \label{alg:CDM}
    	\selectfont
  	    \footnotesize{\textbf{function} CDM($Q, Q_u, Q_l, s_j, \pi $)}
    \begin{algorithmic}[1]
	    \Require 
 	    {$Q$ is the vector of estimated Q-values in the full-space and subspaces, $Q_u$ is the vector of   upper bounds of the Q-values  in the full-space and subspaces, 
 	    $Q_l$ is the vector of  lower bounds of the Q-values in the full-space and subspaces, $s_j$ is the current state of the full-space,  $\pi$ is the policy in the full-space and subspaces}
	    \ForAll{$a \in A$}
	    \State $CI(s_j, a)=[ Q_l(s_j, a), Q_u(s_j, a) ]$
	    \ForAll {$x \in S_j$}
		\State  Calculate $p_\pi(s_x,a)$
	    \State $CI(s_x, a)=[ Q_l(s_x, a), Q_u(s_x, a) ]$
	 \State
	  $CD(s_x, a)=\biggl \{ \begin{array}{ll}
	  f\left(\frac{\vert CI(s_j, a)\vert}{\vert CI(s_x, a)\vert} \right)  \times g \left(\frac{\vert \widehat{Q}_\pi (s_j,a) - \widehat{Q}_\pi (s_x, a)\vert}{\vert CI(s_j, a)\cap CI(s_x, a) \vert}\right), &CI(s_j, a)\cap CI(s_x, a) \neq \emptyset\\ 
	  0,&\text{otherwise}\end{array}$ 
	 \Comment{\parbox[t][]{\dimexpr.25\linewidth-\algorithmicindent}{Choose $f$ and $g$ according to Subsections \ref{SubSection:LengthQuantity} and \ref{SubSection:Overlap-distance-Quantity}.}}
		\EndFor

		\EndFor

		\ForAll {$a \in A$}
	\State $\alpha(s_j,a) = \arg\max_{x \in S_j} CD(s_x,a)$ \Comment{ $S_j$ is the set of subspaces.}
		\State $Pr(s_j, a) = 
\begin{cases}
p_{\pi}(s_{\alpha(s_j,a)},a)   & \quad \text{if } CD(s_{\alpha(s_j,a)},a)>1\\
p_{\pi}(s_j,a) & \quad \text{if } CD(s_{\alpha(s_j,a)},a)\leq1,\\
\end{cases}$
		\EndFor
		\State Choose action $a$ in state $s_j$ with probability $Pr(s_j, a)/\sum_{a}Pr(s_j, a)$\\
		\Return {$a$}
	\end{algorithmic}
\end{algorithm*}
\begin{algorithm*}[!ht]
    \caption{Model based learning with subspaces}
    \newcommand{\NewComment}[1]{ {\hfill$//$ #1}} 
    \label{alg:CIMB}
    \selectfont
   \footnotesize {\textbf{function} MoBLeS($ \pi, \theta, \delta_p, \delta_R$)}    
    \algblockdefx[NAME]{StarT}{EnD}%
	[1]{\textbf{repeat}  #1}
	{\textbf{Ending}}
	\algtext*{EnD}
    \begin{algorithmic}[1]
     \Require 
 	    {$\pi$ is the policy,  $\theta$ is a small positive number for the stopping  criterion used in three recursive algorithms  for computing Q-values (PolicyEvaluation), upper bounds on the Q-values (OptimisticValue), and lower bounds on the Q-values (PessimisticValue), parameters $1-\delta_R \in [0, 1]$ and  $1-\delta_p \in [0, 1]$ are the confidence coefficients.}\\
 	    \footnotesize {\textbf{initialize}\; $Q(.)=0,\ Q_u(.)=0,\ Q_l(.)=0,\ \widehat{R}(.)=0,\ \widehat{p}(.)=\frac{1}{m(.)}$
    }   \Comment{\parbox[t][]{\dimexpr.35\linewidth-\algorithmicindent}{$m(.)$ is the number of neighboring states if available, otherwise $m(.)=|S|$.}}
	 \Statex{}
	 \StarT{for each episode}
	  \State Initialize $s_j$
	  \Repeat{\;in each step of episode}
	  \State	  $a=\text{CDM}(Q, Q_u, Q_l,  s_j, \pi )$ \Comment  Choose action $a$ by using Algorithm \ref{alg:CDM}
	  \State Take action $a$, observe $s_j'$ and $r(s_j, a, s_j')$.
	  \State Update the estimated model in the full-space and in the subspaces by using relations  \ref{equ:AveReward}, \ref{equ:AveProb}, \ref{equ:num_SAS}.
	  \State       $\widehat{Q}_\pi(s_j,a) =\; \widehat{R}(s_j,a)+ \sum_{s_j'}  \widehat{p}(s_j,a,s_j')\left[ \gamma \widehat{V}_\pi(s_j')\right]$ 
	  \Comment{This algorithm for updating Q-values is called full backup~\cite{van2013efficient}}
	 
	  \State Compute an estimate for the confidence interval of $\widehat{Q}_\pi(s_j,a)$ by running lines 5-27 of Algorithm \ref{alg:UB}.
	  \ForAll{$x \in S_j$}
	  \Comment{$S_j$ is the set of  subspaces}
	  \State       $\widehat{Q}_\pi(s_x,a) =\; \widehat{R}(s_x,a)+ \sum_{s_x'}  \widehat{p}(s_x,a,s_x')\left[ \gamma \widehat{V}_\pi(s_x')\right]$ 
	  \State Compute an estimate for the confidence interval of $\widehat{Q}_\pi(s_x,a)$ by running lines 5-27 of Algorithm \ref{alg:UB}. 
	   \EndFor
	  \State $s_j=s_j'$
	  \Until{end of episode}
	  \ForAll{the subspaces and the full-space}
	  \State $Q =\text{PolicyEvaluation}(\pi, \theta)$
	  \State $Q_u =\text{OptimisticValue}(n, \widehat{p},\widehat{R}, \text{LenR}, \delta_R, \delta_p, \gamma, \theta)$ by using Algorithm \ref{alg:UB}
	  \State $Q_l =\text{PessimisticValue}(n, \widehat{p}, \widehat{R}, \text{LenR}, \delta_R, \delta_p, \gamma, \theta)$  by using Algorithm \ref{alg:UB}
	  \EndFor
	  \EnD
    \end{algorithmic}
\end{algorithm*}

\section{Experimental Results}\label{sec:Experimental Results}
We compare the performance of our proposed method (that we call \textit{Model Based Learning with Subspaces}, abbreviated as MoBLeS) with four other off-policy approaches: 
\begin{table}[H]
\normalsize
\begin{tabular}{p{1cm}p{6.5cm}}
MB & \textit{Model Based} learning as explained in the background. It is the same as our method without considering subspaces.\\
Q($\lambda$)& Famous \textit{Q($\lambda$)} learning method~\cite{watkins1989learning} by selecting the best decay rate of the eligibility trace ($\lambda$) and learning rate ($\alpha$).\\
QS($\lambda$)& \textit{Q($\lambda$)} with subspaces is a modified version of~\cite{daee2014}. The algorithm in~\cite{daee2014} uses \textit{Monte Carlo} approach which is very slow. Hence, we use \textit{Q($\lambda$)} learning with the best $\lambda$ and $\alpha$ while uncertainties are calculated according to the \textit{Monte Carlo} as explained in~\cite{daee2014}. \\
QL & \color{black}The model-free learning method that uses a linear function approximator with tile coding~\cite{sutton2016reinforcement}. Here, each subspace is considered as a tiling. QL algorithm has a learning rate ($\beta$) for learning the function approximator. 
\end{tabular}
\end{table}

The policy of the agent in both the full-space and subspaces is \textit{$\epsilon$-greedy} with the same \textit{$\epsilon$}. The parameters needed  for the learning process are given in Tables \ref{T:Learningـparameter}-\ref{T:alpha}. All models need $\gamma$ and $\epsilon$ parameters. The model-based approaches, meaning MB and MoBLeS need $\theta$ parameter. MoBLeS and QS($\lambda$) require $\delta_R$. MoBLeS needs $\delta_p$. In the simulations for Q($\lambda$) and QS($\lambda$), we consider different learning parameters according to Tables~\ref{T:lambda},~\ref{T:alpha} and the best results are reported in the comparisons. {\color{black} We tried different learning rates as shown in Table~\ref{T:alphaQL} and report the best results in the comparisons.}
\begin{table}
    \small
    \selectfont\centering
    \begin{center}
	\caption{Values of the required parameters for algorithms}
	\label{T:Learningـparameter}
	\begin{tabular}{|c|c|p{4cm}|}
	    \hline	
	    \textbf{Parameter} 	&	\textbf{Value}	&	 \textbf{Parameter description}\\
			    \hline 	
			    \hline			
	    $\gamma$ 	&	$0.9$	&	 Discount factor\\
			    \hline	
	    $\epsilon$ 	&	$0.1$	&	 $\epsilon$ in the $\epsilon-greedy$ policy\\
			    \hline		
	    $\theta$ 	&	$\frac{0.01}{1 + \log \left(\text{\#episode}\right)}$	& Stopping criterion of the PE algorithm and the algorithms for computing the optimistic and pessimistic Q-values\\
			    \hline
	    $1-\delta_R$ 	&	$0.9$	&Confidence coefficient of the estimated rewards in the full-space and subspaces\\
			    \hline		
	    $1-\delta_p$ 	&	$0.9$	&Confidence coefficient of the estimated transition probabilities in the full-space and subspaces\\
			    \hline			
	\end{tabular} 
    \end{center}
\end{table}
\begin{table}
    \small
      \parbox{.45\linewidth}{
	\begin{center}
	\caption{\mbox{Values of $\lambda$}}
	\label{T:lambda}
	\begin{tabular}{|c|c|}
	    \hline
	    \textbf{Number}	&	\textbf{Value}
	    \\		\hline 
			    \hline
	    1	&	$1$
	    \\		\hline
	    2	&	$0.9$
	    \\		\hline
	    3	&	$0.5$
		    \\	\hline
	    4	&	$0$
	    \\		\hline		
	\end{tabular} 
      \end{center}
      }
      \hfill
      \parbox{.55\linewidth}{
	 \begin{center}
	\caption{Values of $\alpha$ for Q($\lambda$) and QS($\lambda$)}
	\label{T:alpha}
	\begin{tabular}{|c|c|}
	    \hline
	    \textbf{Number}	&	\textbf{Value}
	    \\		\hline 
			    \hline
	    1	&	$\frac{1}{1+ n(s, a)}$
	    \\		\hline
	    2	&	$\sqrt{\frac{1}{1+n(s,a)}}$
	    \\		\hline
	    3	&	$\frac{1}{1+\sqrt{n(s, a)}}$
	    \\		\hline
	    4	&	$\frac{1}{1+\text{\# episodes}}$
	    \\		\hline
	    5	&	$\sqrt{\frac{1}{1+ \text{\# episode}}}$
	    \\		\hline
	    6	&	$\frac{1}{1+\sqrt{\text{\# episode}}}$
		    \\	\hline
	    7	&	$0.1$
	    \\		\hline
	\end{tabular} 
      \end{center}
      }
       \begin{center}
       	\caption{Values of $\beta$ for QL}
       	\label{T:alphaQL}
       	\begin{tabular}{|c|c|}
       		\hline
       		\textbf{Number}	&	\textbf{Value}
       		\\		\hline 
       		\hline
       		1	&	$\frac{1}{1+\text{\# subspaces}}$
       		\\		\hline
       		2	&	$\frac{1}{1+ 10\times\text{\# subspaces}}$
       		\\		\hline
       		3	&	$\frac{1}{1+\text{\# episodes}}$
       		\\		\hline
       		4	&	$\sqrt{\frac{1}{1+\text{ \# episode}}}$
       		\\		\hline
       		5	&	$\frac{1}{1+\sqrt{\text{\# episode}}}$
       		\\	\hline
       	\end{tabular} 
       \end{center}
\end{table}

\begin{table*} 
  \small
  \begin{center}
    \caption{Distribution of the rewards for different scenarios in the 2D maze task.}
    \label{T:Envparameter}
    \begin{tabular}{|l|l|}
	    \hline		
	    \textbf{Parameter description} 	&	\textbf{Value}	
	    \\		\hline  \hline		
	    The immediate reward of collision with the wall  	&	$\sim \frac{1}{3} \mathcal{N}(-11.5, 0.2) + \frac{2}{3} \mathcal{N}(-10.5, 0.3), \; \in [-12, -10]$ 	
		    \\	 \hline		
	    The immediate reward of attaining the goal	&	$\sim N(+10, 0.02), \; \in [9.5, 11.5]$	
	    \\		 \hline
	      The immediate reward of each step 	&	$\sim \frac{1}{3} \mathcal{N}(-1.5, 0.2) + \frac{2}{3} \mathcal{N}(-0.5, 0.3), \; \in [-2, 0]$	
	    \\		\hline		
      \end{tabular} 
  \end{center}
\end{table*}

The measure for the comparison is the average accumulated reward received by the agent in each episode of the learning. We run each experiment 10 different times with the same starting point. The error-bars and the average measures are shown on the plots. We stop the learning process after 100 episodes of the learning.

In the simulation studies, we considered two set of experiments, a set with two sensors (features) and another with six sensors.
The environments in both experiments are 2D mazes with some barriers (see Fig.~\ref{fig:2D-Environments}). The agent learns the task of reaching a goal from a certain initial point. The task is finding the average shortest path from starting point to the end point. 
The goal is fixed but starting point in the beginning of each episode is selected by random. For  environments with two features, feature set is selected to be the location of the agent $F=\lbrace x, y \rbrace$. For experiments with six features, we add four extra infrared  sensors in addition to $x$ and $y$ positions of the agent. We will later explain the meaning of the extra infrared sensor.

The state of the agent in the environments is the position of the agent. In each state, four actions are available to the agent: up, down, right and left. 
When the agent chooses an action, it moves on a random direction with probability $P_{\text{rand}}=0.1$
and it moves along the intended direction with probability $1-P_{\text{rand}}$. If there is a barrier in the new position, the agent remains in the current location. The immediate rewards obtained by each agent's action are shown in the Table~\ref{T:Envparameter}. The rewards come from truncated mixtures of Gaussian distributions.
Only the knowledge of the interval lengths of rewards and the number of states reachable through each action is used in our proposed method MoBLeS.

In the simulations of the model-based approaches, the initial value for the probability of the transitions is set to be equal. The initial values of the estimated rewards are set to zero. As we explained previously, because PE algorithm is expensive it runs only after each episode. During the episode, the Q-values are updated by applying only one step of Bellman equations given in~\eqref{equ:Q_Aprox} and~\eqref{equ:V_Aprox} similar to Dyna-Q learning algorithm~\cite{sutton1990integrated} and small backups algorithm~\cite{van2013efficient}.
 {\color{black}Full backup within each episode provides a significant performance benefit. Without full backup, the learning is deferred to the end of each episode. This results in delaying updating policies and therefore slower learning.}

\subsection{Simulations with 2 sensors}
\begin{figure}
      \centering
      \begin{subfigure}{0.45\columnwidth}
	      \includegraphics[width=1\columnwidth]{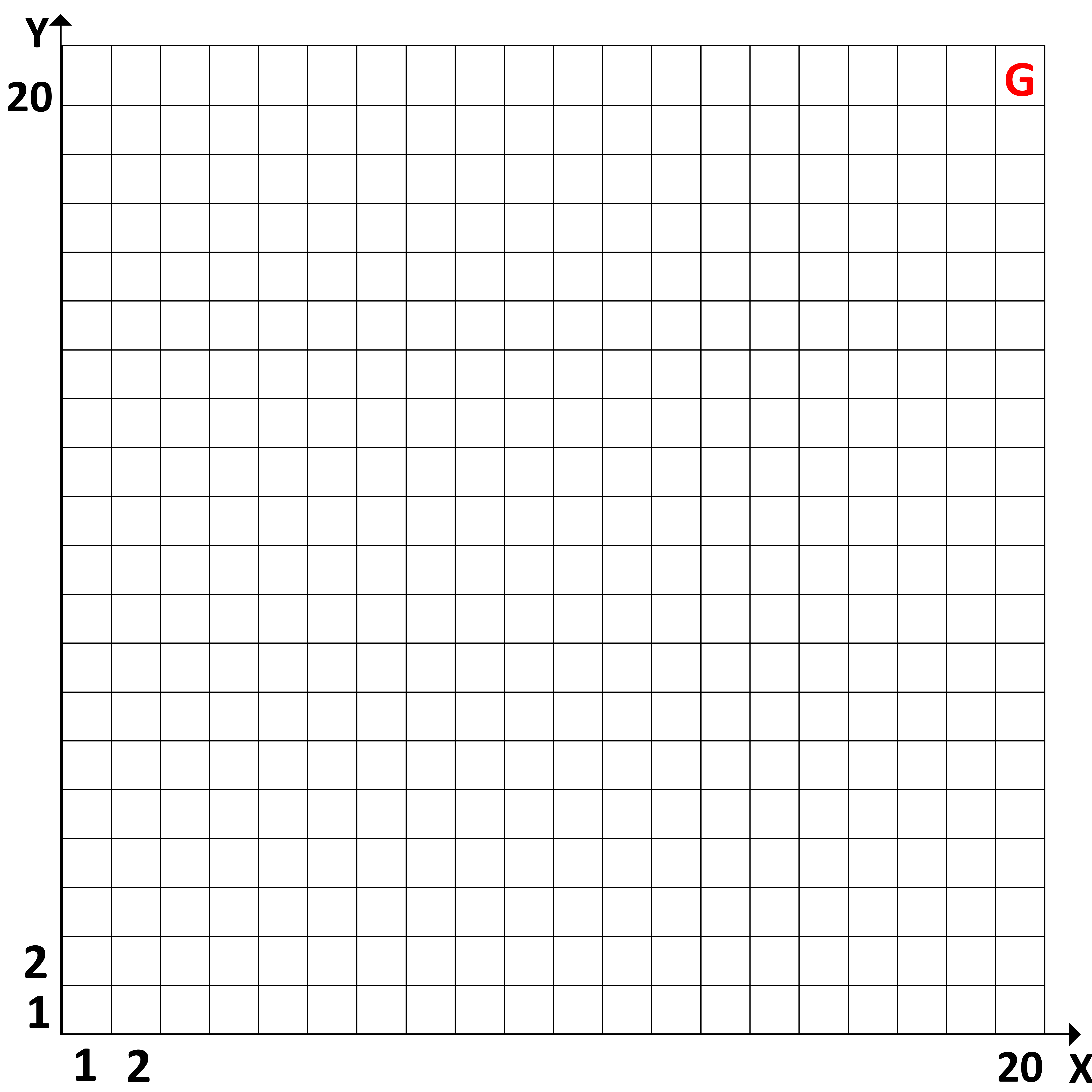}
	      \caption[Environment 1: Big environment without any obstacle]
	      {Environment 1: No obstacle}
	   
	    \label{fig:VeryBigEnv}
    \end{subfigure}
     ~
    \begin{subfigure}{0.45\columnwidth}
	      \includegraphics[width=1\columnwidth]{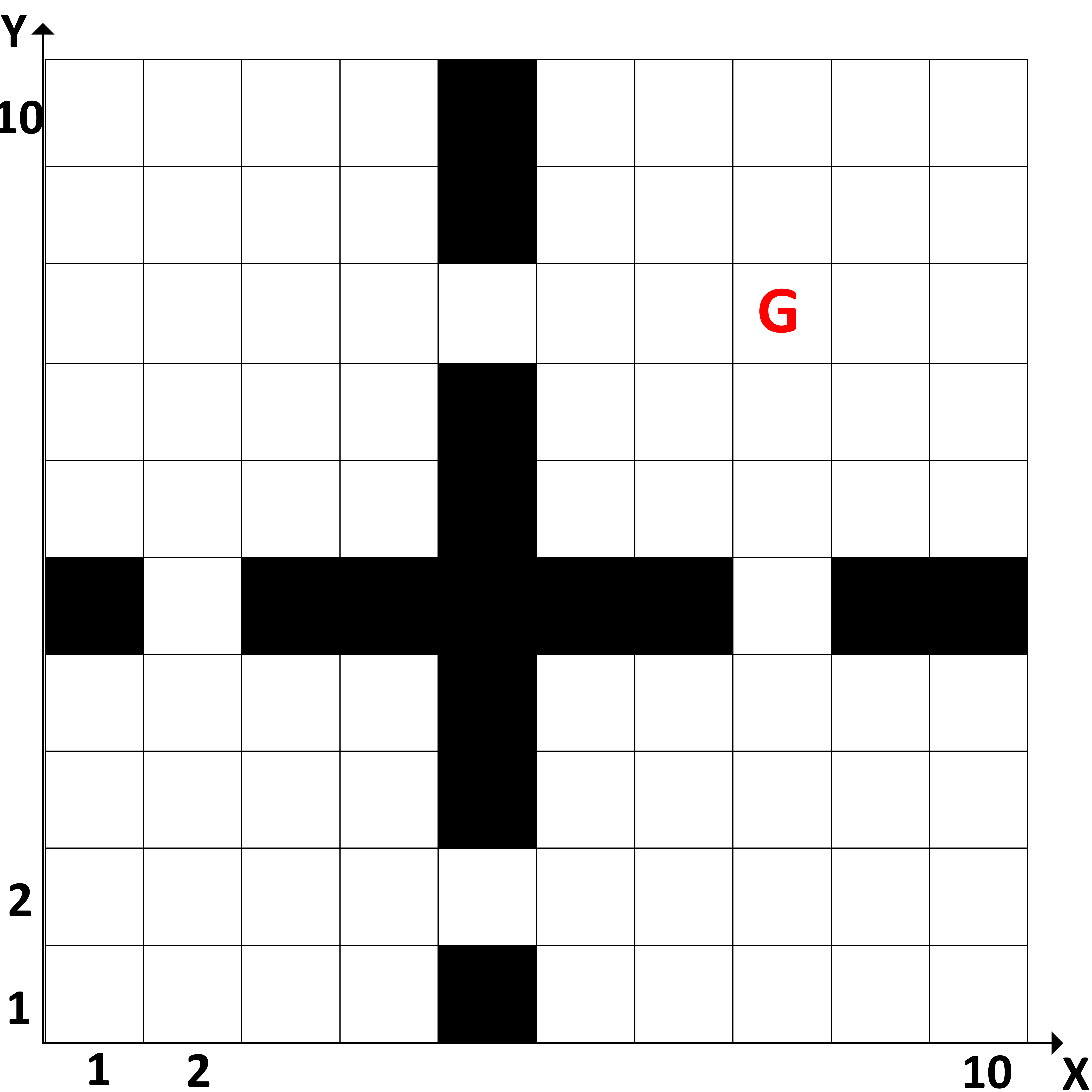}
	       \caption[Environment 2: Four-rooms]
	      {Environment 2: Four-rooms}
		
	    \label{fig:4Rooms}
      \end{subfigure}    
      \medskip
      
      \begin{subfigure}{0.45\columnwidth}
	      \includegraphics[width=1\columnwidth]{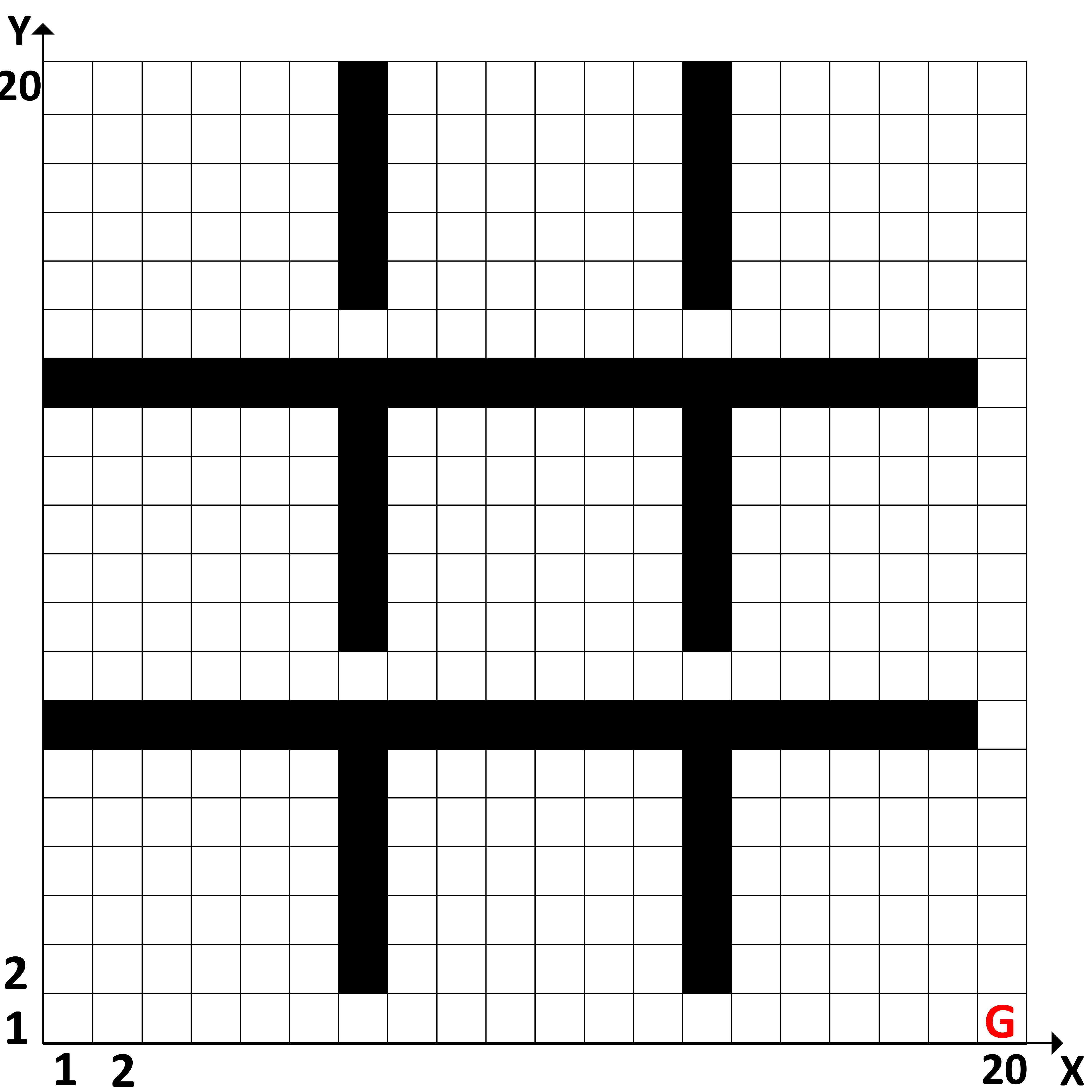}
	      \caption[Environment 3: Nine-rooms]
	    {Environment 3: Nine-rooms}
	      \label{fig:9Rooms}
      \end{subfigure} 
	~
	\begin{subfigure}{0.45\columnwidth}
	      \includegraphics[width=1\columnwidth]{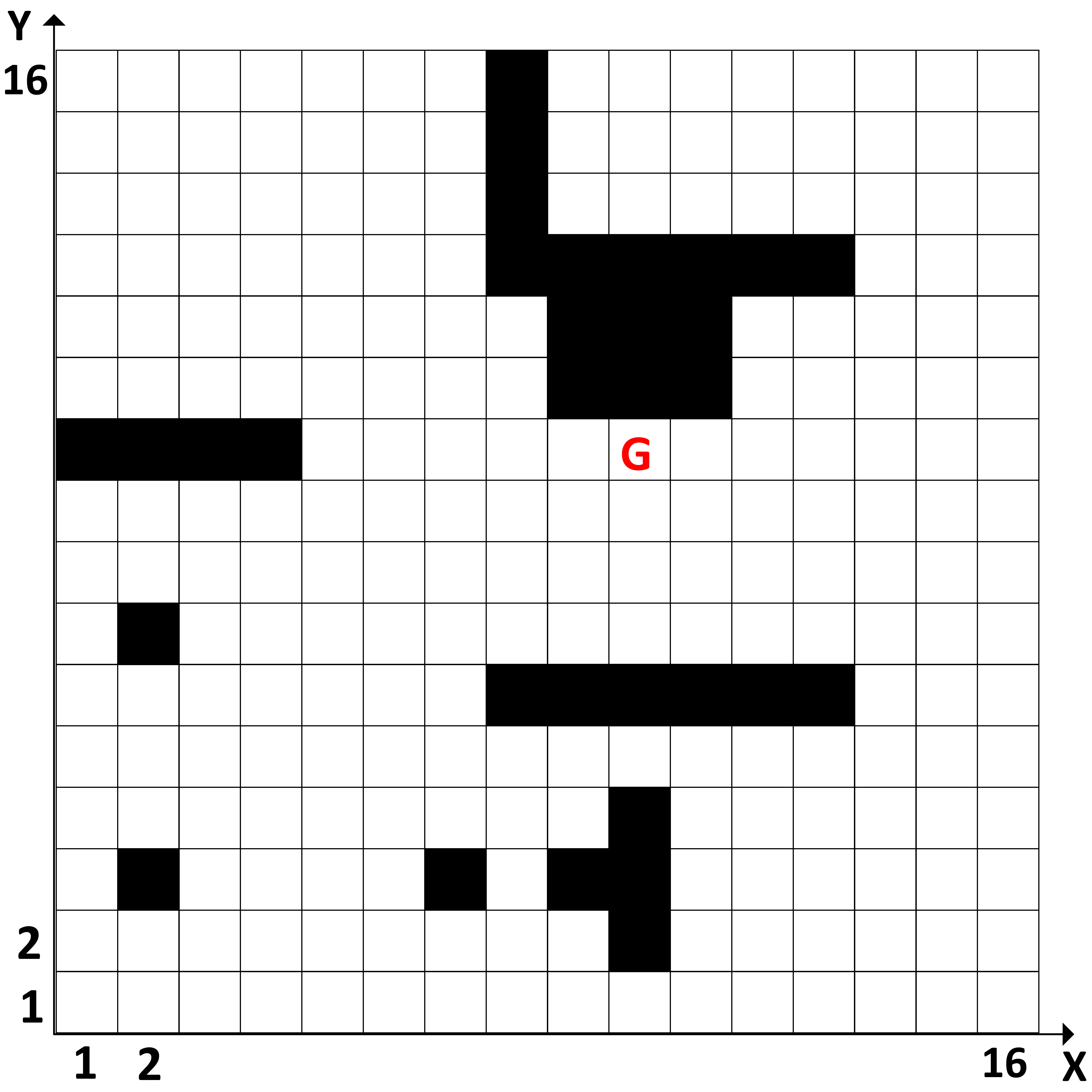}
	      \caption[]
    {Environment 4: Semi-random obstacles}
	    \label{fig:BigEnvOBS}
      \end{subfigure} 
  \caption[2D-Environments]
  {Four different mazes used in our simulated studies. The goal is shown by ``G'' in the environments and the task is to find the goal in the fewest number of steps. Each episode starts in a random state of an environment.
}
  \label{fig:2D-Environments}
\end{figure}

Four different environments used in our simulations for 2 features are shown in Fig.~\ref{fig:2D-Environments}. The first environment has no
obstacle. The second and third environments have room-like structure. In the fourth environment, the obstacles are arranged semi-randomly. These environments are chosen to be diverse so to see how subspaces may help in different scenarios.  

\begin{figure*}
\vspace{-2.3cm}
	\centerline{
		\resizebox{1\textwidth}{!}{\input{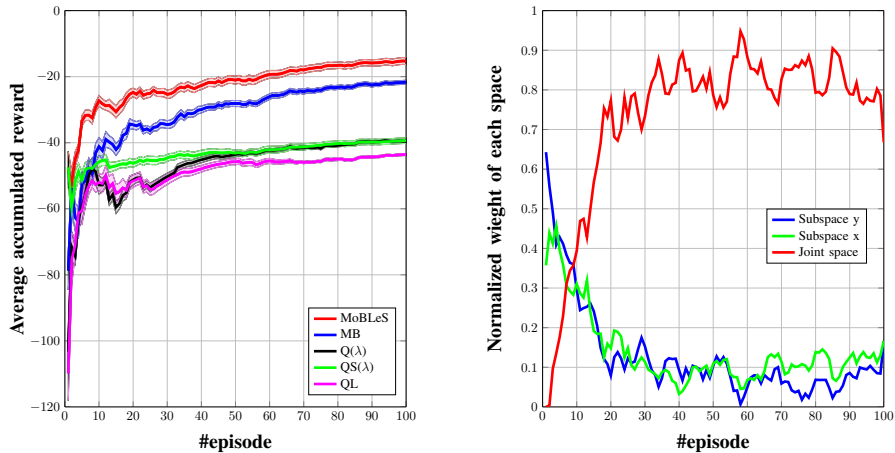}}
	}
	\caption{ Result of the environment with no obstacle. Left plot shows average accumulated reward as a function of the number of  episodes averaged over 15 different runs. Shaded area shows standard error of the mean. Right plot shows the mean weights of the subspaces and the full-space for MoBLeS method. The weights are smoothed using the rectangular smoothing window of length 5. }
	\label{fig:EnvOne}    
\end{figure*}

Figures~\ref{fig:EnvOne}-\ref{fig:All} show the simulation results for different environments shown in Fig.~\ref{fig:2D-Environments}. Left plot in Fig.~\ref{fig:EnvOne} and the plots on the first row of Fig.~\ref{fig:All} show the average accumulated rewards as a function of the number of episodes averaged over 15 different runs. Other plots show the weights of the subspaces and the full-space computed in the proposed confidence degree model averaged over all steps in each episode.  Large weights for the subspaces show that the agent rely on their decisions. In the beginning episodes of the learning, the weights of the subspaces are large because of the generalization of them. In simpler environments where the subspaces are more informative, the weights decay slowly while for harder environment it decays fast. That is because of the problem of the PA issue in the subspaces. 

\begin{figure*}	
    \centerline{
    		\resizebox{1\textwidth}{!}{\input{Pic/Result/EnvNoSen.tex}}
    }
\caption{The columns from left to right show the results of the four-rooms environment, the nine-rooms environment and the environment with semi-random obstacles with two features. The upper row shows average accumulated reward and the lower row shows mean weights of the subspaces and the full-space for MoBLeS method. The description of the plots are the same as Fig.~\ref{fig:EnvOne} }
\label{fig:All}
\end{figure*}

\begin{figure*}	
    \centerline{
    		\resizebox{1\textwidth}{!}{\input{Pic/Result/EnvExtraSen.tex}}
    }
\caption{The columns from left to right show the results of the four-rooms environment, the nine-rooms environment and the environment with semi-random obstacles with six features. The upper row shows average accumulated reward and the lower row shows mean weights of the subspaces and the full-space for MoBLeS method with extra IR sensors. The description of the plots are the same as Fig.~\ref{fig:EnvOne}. }
\label{fig:irAll}
\end{figure*}

 A clear example of the effect of the PA issue on the weight decay is the difference of weight decay between the second and the third environments. Since in the third environment, doors between rooms are aligned along directions $x$ and $y$, the subspaces generalize better and the problem of the PA issue is less severe. Therefore, the weights decay slower in the third environment. In the fourth environment with difficult semi-random obstacles, the weights decay very fast, but even in this case, using subspaces do not deteriorate the performance of our proposed method.

The method MoBLeS-Thr in the last column of Fig.~\ref{fig:All} corresponds to our method where after the fifth visit of each state-action during the learning we put the weights to zero meaning that we switch to the full-space mode. It can be seen that this threshold helps in improving the performance of our method. This shows that our formulas for computing CDs of the subspaces can be further improved.

 In all environments, the proposed method reaches higher accumulated reward compared to the MB method. It can be seen that the difference is higher in the early episodes of the learning process.

\subsection{Simulations with 6 sensors}

In this part, we investigate the effect of adding extra sensors to the 2D environments of the previous simulation. The environments are shown in the figures ~\ref{fig:4Rooms}, \ref{fig:9Rooms}, and  \ref{fig:BigEnvOBS}.  
The extra sensors are four infrared (IR) ones that each shows the presence of a barrier in front of a direction. These directions in our experiments are up, down, right and left. Each sensor can take two values, either one or zero. Value one shows the presence of a barrier in the direction of that sensor and zero means there is no barrier. Therefore, the agent has the total number of six sensors, two for its main sensors $\{x, y\}$ and four for IR sensors \{IR-up, IR-right, IR-down, IR-left\}.

The IR sensors convey extra information about the environment. The agent can learn the MDP task without these extra sensors, but incorporating these extra sensors may expedite the learning process. When using MB learning for the full-space, these extra sensors are redundant and the agent can not exploit the information conveyed by these sensors. However, We can simply assume each extra sensor to be a subspace and use our procedure for learning the MDP task. 

Figure~\ref{fig:irAll} shows the result of the proposed method with 6 sensors in compare to 2 sensors, MB learning, and QL with 6 tiling. 
We can see that in all figures, having extra sensors increases the speed of learning in the early stages. It is because  the IR sensors have more generalization than $x$ and $y$ sensors, i.e. the state-space of the IR sensors is smaller and the agent learns faster in them. In the environment of Fig.~\ref{fig:9Rooms}, where this extra information is more helpful, we observe significant improvement of MoBLeS with 6 sensors.

\section{Conclusion}
In this paper, we introduced a new methodology, called MoBLeS, for improving the learning speed of the model-based approaches in the reinforcement learning. We assume that the agent does not have any prior knowledge about the task. Our method attacks the curse of dimensionality problem by defining several MDPs for different subspaces of the full-space in addition to the full-space MDP and then combining the decisions of these MDPs. 
{\color{black}Here, the curse of dimensionality means a large amount of  experiences required to learn the task.} Our methodology of combining the decisions balances the opposite effects of generalization and perceptual aliasing of the subspaces. Subspaces have better generalization, it means that the learning process can be faster in them, because they are lower-dimensional. However, they have the problem of perceptual aliasing, because they are partial observable and this makes their decisions in some cases sub-optimal. 

For each state-action pair, the best MDP is selected based on its confidence degree. The confidence degree shows the reliability of each space decision. Our proposed method for computing confidence degrees was 
based on the  estimated Q-values and their confidence intervals. The generalization and PA issue can be seen by comparing the estimated Q-values and their confidence intervals of a subspace and the full-space as explained in the text.

The number of the parameters of the model is equal to $M|S||A|+|S||A|$, where $|S|$ is the number of states, $M$ is defined as $ \max_ {s \in S} m(s)$ and $|A|$ is the number of actions. The number of the parameters in each subspace is the same with a difference that $S$ and $M$ are replaced by that of each subspace. Therefore, if the dimensionality of each subspace is significantly smaller than the full-space, the number of the parameters and subsequently the computational complexity are also smaller. {\color{black}Therefore, the computational complexity of our method is  slightly higher than the model-based approach and we use policy iteration, a computationally intensive method.}

Convergence of the method to the optimal policy was proven mathematically. Furthermore, we tested its performance on some mazes ranging from simple to complex ones. The results showed improvement in the learning speed especially in the early stages of the learning process. Improving the learning speed at the early trials is of significant practical importance, especially when the agent has limited learning budget. In the environments where the subspaces are informative, i.e. the generalization in the subspaces are considerable, we saw significant speed-up using our method in the beginning episodes. Even in the complex environments, our method works reasonably well and is not defeated by the pure model-based method in the full-space.

One interesting feature of our proposed method is that it can easily incorporate extra sensors in the learning. 
From the point of the state-space, these sensors are redundant and therefore they are not exploited in the original model-based method. However, in MoBLeS we benefit from the subspaces formed by the extra sensors to increase generalization and increase the learning speed. In our simulations in the environments with extra infrared sensors, we saw significant improvement in the learning speed.

There are multiple directions for future works:
\begin{itemize}
\item We saw that our method works particularly well in environments where subspaces are more informative. To this end, for more difficult environments, it is better to split the environment into several sub-environments with some subgoals, it is called skill acquisition in the literature \cite{mehta2008automatic, barry2011deth, taghizadeh2013novel}. Then, our proposed method can be used in each sub-environment with informative subspaces.
\item Our framework can be developed for partial observable MDPs. We think that the method will work very well in these scenarios. For example, consider that some sensors have high noise. These sensors are less reliable and these sensors can be neglected in the planning.
\item Not all subspaces are informative and therefore one may design a procedure for removing some uninformative subspaces during the learning. This procedure also reduces the computational complexity.
\item The procedure for combining the decision making in the subspaces can be further improved. For example in Fig.~\ref{fig:BigEnvOBS}, we saw that including threshold improves the learning speed. {\color{black} One can use a parametric model like neural networks for the CDM and train the parameters on some sample tasks. The objective would be to find the parameters minimizing the averaged regret on those tasks.}
\item {\color{black}The method can be extended to huge discrete state-spaces and continuous spaces. A promising direction is using deep learning approach~\cite{oh2015action}.}
\end{itemize}

\bibliographystyle{IEEEtran}
\bibliography{MyRef}

\end{document}